\documentclass{article}

\usepackage[preprint, nonatbib]{neurips_2019}

\usepackage[utf8]{inputenc} 
\usepackage[T1]{fontenc}    
\usepackage{hyperref}       
\usepackage{url}            
\usepackage{booktabs}       
\usepackage{amsfonts}       
\usepackage{nicefrac}       
\usepackage{microtype}      

\usepackage{amsmath}
\usepackage{caption}
\usepackage{gensymb}
\usepackage{biblatex}
\usepackage{hyperref}
\hypersetup{
    colorlinks=true,
    linkcolor=blue,
    filecolor=blue,      
    urlcolor=blue,
}
\usepackage{bm}

\def\eqref#1{equation~\ref{#1}}

\def\1{\bm{1}}

\def\vtheta{{\bm{\theta}}}

\def\vc{{\bm{c}}}

\def\vx{{\bm{x}}}
\def\vy{{\bm{y}}}
\def\vz{{\bm{z}}}

\def\mM{{\bm{M}}}

\DeclareMathAlphabet{\mathsfit}{\encodingdefault}{\sfdefault}{m}{sl}
\SetMathAlphabet{\mathsfit}{bold}{\encodingdefault}{\sfdefault}{bx}{n}

\newcommand{\anon}[1]{#1}
\newcommand{\hide}[1]{#1}

\newcommand{\loss}{\ell}

\title{Ward2ICU: A Vital Signs Dataset of Inpatients from the General Ward}

\author{%
  Daniel Severo\\
  3778 Healthcare\\
  São Paulo, Brazil\\
  \texttt{severo@3778.care} \\
   \And
   Flávio Amaro \\
   3778 Healthcare \\
   Belo Horizonte, Brazil \\
   \texttt{flavio@3778.care} \\
   \AND
   Estevam R. Hruschka Jr \\
   Carnegie Mellon University \\
   Pittsburgh, USA \\
   \texttt{estevam@cs.cmu.edu} \\
   \And
   André Soares de Moura Costa \\
   Mater Dei Healthcare \\
   Belo Horizonte, Brazil \\
   \texttt{andre.costa@materdei.com.br} \\
}

\begin{document}

\maketitle

\begin{abstract}
    We present a proxy dataset of vital signs with class labels indicating patient transitions from the ward to intensive care units called \emph{Ward2ICU}. Patient privacy is protected using a Wasserstein Generative Adversarial Network to implicitly learn an approximation of the data distribution, allowing us to sample synthetic data. The quality of data generation is assessed directly on the binary classification task by comparing specificity and sensitivity of an LSTM classifier on proxy and original datasets. We initialize a discussion of unintentionally disclosing commercial sensitive information and propose a solution for a special case through class label balancing.
\end{abstract}

\section{Introduction}

Public datasets are a crucial component for the advancement of science \cite{baxevanis2015importance}. Acquiring labeled data is essential to Machine Learning tasks and often very expensive. These datasets allow for a common ground for comparison between different algorithms and models. Techniques such as transfer learning can be used to lift performance on tasks not originally associated with the published dataset \cite{yosinski2014transferable}. For example, pre-training on ImageNet \cite{deng2009imagenet} for computer-vision tasks is now a common practice. Healthcare is no different, but concerns with patient privacy and commercial sensitive information hinder the publication and dissemination of datasets by institutions \cite{kostkova2016owns}. The Machine Learning community has benefited significantly from datasets such as MNIST \cite{lecun1998gradient}, ImageNet and WordNet \cite{miller1998wordnet}, but there are few widespread databases that lead to well defined machine learning tasks in health and bioinformatics such as MIMIC \cite{johnson2016mimic} and eICU \cite{pollard2018eicu}.

\paragraph{Issues beyond patient privacy} Guaranteeing patient privacy is an ongoing field of study \cite{beaulieu2019privacy,walsh2018enabling}. The possibility of unintentionally revealing commercial sensitive information is also a great obstacle for the availability of public datasets and is generally not discussed. For example, if a hospital's occupancy rate can be inferred by a health insurance company it can be used as leverage during negotiations. Competitors may use patient population statistics derived from clinical datasets for targeted commercial campaigns in an attempt to gain market share. Healthcare providers are also reluctant to disclose the specifics of their care practices, concerned that it may be used for benchmarks by competitors.

\paragraph{Our contributions}
\begin{enumerate}
    \item Release a new anonymized vital signs dataset inducing a binary classification task of patient transitions from the general ward to an intensive care unit called \emph{Ward2ICU};
    \item Discuss the aforementioned issue of hiding commercial sensitive data and demonstrate a possible solution in our context.
\end{enumerate}

Although vital signs are not considered as sensitive as other patient data (e.g. exam results, age, gender), we create a proxy dataset using a Conditional WGAN-GP to mitigate privacy concerns \cite{gulrajani2017improved}. A classifier that shows similar performance when trained on the proxy and original datasets is built using LSTM \cite{gers1999learning} and a fully connected layer.

\begin{table}
    \centering
    \captionsetup{skip=0.5\baselineskip}
    \caption{Lower and upper bounds of vital signs filters.}
    \label{vs-ranges}
    \begin{tabular}{llll}
         \toprule
        Vital Signs                       & Unit        & Lower & Upper\\ \midrule
        Temperature                       & \degree C   & 30    & 45 \\
        Respiratory Rate                  & breaths/min & 5     & 75 \\
        Heart Rate                        & beats/min   & 10    & 250 \\
        Systolic Arterial Blood Pressure  & mmHg        & 20    & 300 \\
        Diastolic Arterial Blood Pressure & mmHg        & 10    & 200 \\
        \bottomrule
    \end{tabular}
\end{table}

For experiments, we used TorchGAN \cite{pal2019torchgan}, GNU Parallel \cite{Tange2011a} and our own source code which has been made available together with a synthetic pre-release of our dataset. \footnote{\anon{\url{https://research.3778.care/publication/ward2icu}}} Our long term goal is to progressively publish other datasets after surveying the research community to direct our efforts. \footnote{\anon{\url{https://research.3778.care/publication/survey}}}




\section{Original Dataset}

\emph{Ward2ICU} is a dataset of sequential physiological measurements regarding the vital signs discussed below together with a a binary class label. It derives from Electronic Health Records (EHR) of patients from \anon{\emph{Hospital Mater Dei} (HMD)}, a tertiary hospital, located in \anon{Belo Horizonte, Brazil}. It consists of adult patients with an average age of 40, admitted to the standard ward between the years of 2014 and 2019. Over 25 vital signs are monitored and collected but only 5 have been made available as of the present date. Each data point was measured and recorded manually by nursing professionals. The default interval between measurements is 6 hours but this is sometimes overlooked when demanded by medical staff. This results in an average of 4 to 4.6 data points for each of the 5 different vital signs taken per day per patient. We define a \emph{sample} as the measurement of all 5 vital signs near simultaneously for a single patient. For each patient, 20 sequential samples are provided totaling 100 data points, 20 for each vital sign. A filtering stage removes patients that have \emph{at least one} sample outside the pre-defined ranges shown in \autoref{vs-ranges}. Patients with label $1$ have been moved to the ICU by the time the 21st sample is taken while $0$ indicates a discharge. The class ratio is a commercial sensitive information to \anon{HMD}, hence the exact number can not be disclosed. However, we can confirm that ICU transitions (i.e. the minority class) lie between 5 to 30\%.


\paragraph{Body Temperature (T)}

The average human body temperature ranges from 36.5 to 37.5 \degree Celsius, or 97.7 to 99.5 \degree Fahrenheit
\cite{hutchison2008hypothermia}.
It was routinely measured using a digital thermometer inserted into the mouth, anus, or placed under the armpit.
\paragraph{Respiratory Rate (RR)} The average number of breaths taken per minute. This rate varies depending on the age range. An adult’s normal respiration rate at rest is 12 to 20 breaths per minute \cite{barrett2009ganong}. RR was measured by looking at the patient's chest movements and counting the number of cycles of inhalation and exhalation (i.e. the rise and the fall of the chest wall) per minute \cite{lindh2013delmar}.

\paragraph{Heart Rate or Pulse Rate (HR)}
The number of heart beats over a period of 60 seconds. This vital sign was measured by touching the lateral area of the wrist using the finger tips, where an artery passes close to the surface of an underlying bone. This is a commonly executed maneuver \cite{lindh2013delmar}. We also used a digital pulse oximeter to measure the heart rate. It consists of a small display and a sensor attached to the patient's finger that measures and displays the data \cite{lindh2013delmar}.

\paragraph{Arterial Blood Pressure (ABP)}
The cardiac cycle consists of the events (i.e. diastole and systole) that occur from the beginning of one heartbeat to the beginning of the next. We measured ABP by indirect means, using the Auscultatory method. It consists of inflating a manometer cuff around a patient's arm and listening with a stethoscope for specific sounds that mark the levels of systolic and diastolic blood pressures.



\begin{table}
    \centering
    \captionsetup{skip=0.5\baselineskip}
    \caption{Accuracy on binary classification task.}
    \label{results}
    \begin{tabular}{lllll}
        \toprule
        & \multicolumn{2}{l}{Discharge $(y=0)$} &\multicolumn{2}{l}{ICU $(y=1)$} \\
        \cmidrule(r){2-5}
        Vital Signs    & Real  & Proxy & Real  & Proxy \\ \midrule
        T              & 0.441 & 0.373 & 0.703 & 0.761 \\
        T, RR          & 0.601 & 0.624 & 0.612 & 0.590 \\
        T, RR, HR      & 0.494 & 0.488 & 0.672 & 0.785 \\
        \textbf{T, RR, HR, ABP} & \textbf{0.732}  & \textbf{0.721}  & \textbf{0.478}  & \textbf{0.380}  \\
        \bottomrule
    \end{tabular}
\end{table}

\section{Related work}
Recent work in protecting patient privacy has made significant use of Generative Adversarial Networks (GAN) for synthesizing proxy datasets \cite{goodfellow2014generative}. GAN are a family of generative models for implicit density estimation where a Generator ($G$) and Discriminator ($D$) are trained simultaneously in a zero-sum game. Given the underlying data distribution $p$, $D$'s objective is to classify incoming samples $\vx$ as being real ($\vx \sim p$) or fake ($\vx \sim q$). Meanwhile, $G$ learns to generate samples that can fool $D$ into classifying them as real, by minimizing $d(p, q)$ for some distance $d$. $G$ has no direct access to $p$ and learns only from the gradient signal provided by $D$ together with some loss function $\loss$ induced by $d$. By varying $D$, $G$ and $d$ we can recover most GAN variants. For example, \cite{goodfellow2014generative} uses the Jensen-Shannon Divergence (JSD) \cite{lin1991divergence} for $d$ while Least Squares GAN uses Pearson $\chi^2$ \cite{mao2017least}. Deep Convolutional and Recurrent GAN both minimize JSD, but differ by using convolutional and recurrent architectures, respectively \cite{radford2015unsupervised,esteban2017real}. Wasserstein GAN (WGAN) minimize the Earth mover's distance (EMD), also called Kantorovich–Rubinstein or Wasserstein metric \cite{arjovsky2017wasserstein}. Intuitively, the EMD between $p$ and $q$ is the minimal effort required to transform $p$ into $q$ by transporting density values $p(\vx)$ to $q(\vy)$, or vice-versa. Training JSD GAN suffers from an issue called mode collapse, where the fake samples generated have low diversity (e.g. on MNIST fake images would all be of the same digit). Theoretical and empirical results are given in \cite{arjovsky2017wasserstein}, showing that WGAN have better convergence properties than JSD GAN due to non-vanishing gradients. Some authors have made extensions to include conditional information $p\left(\vx \mid \vc\right)$ such as class labels during learning, usually through the use of embeddings concatenated with input and hidden layers \cite{mirza2014conditional, esteban2017real}. This augmentation can, in theory, be applied to any $(D,G,d)$ triplet.

In practice, training is done by minimizing $\loss$, while $D$ and $G$ are neural networks with parameters $\vtheta$. $G$ transforms some seed distribution $\phi$ into $q$ by sampling values $\vz \sim \phi$ such that $G_\vtheta(\vz) = \vx \sim q_\vtheta$. Currently, there is no $(D,G,d)$ combination that reaches state-of-the-art on all synthesis tasks as it is highly sensitive to the domain of $p$ (i.e. the data type). Previous work in generating synthetic datasets for healthcare and bioinformatics has been done for count \cite{baowaly2018synthesizing, walonoski2017synthea}, binary \cite{baowaly2018synthesizing, walonoski2017synthea}, categorical \cite{choi2017generating}, time-series \cite{esteban2017real,hartmann2018eeg,abdelfattah2018augmenting,harada2019biosignal}, text \cite{guan2018generation} and image data \cite{nie2017medical}.






\section{Synthesis}
To generate a proxy dataset for \emph{Ward2ICU} we employed a Conditional WGAN-GP \cite{gulrajani2017improved}. A one dimensional convolutional network was used for both $D$ and $G$ similar to \cite{hartmann2018eeg} and can be seen in \autoref{gan-network}. RMSprop \cite{tieleman2012lecture} optimizer was used for both networks with default recommended values. The input data was scaled to the interval $[-1, 1]$ by subtracting the mean and dividing by the maximum absolute value \emph{along the channel dimension}. For each epoch, we sampled with repetition a mini-batch keeping the classes uniformly distributed. 30\% of the original data was held out for testing.

The proxy datasets were made to have the same size as the original but with balanced classes. Synthesis quality is evaluated by training a classifier composed of an LSTM with a fully connected layer and computing accuracy for both classes. We varied the total number of vital signs used throughout the experiments. To obtain the results in \autoref{results}, we did a randomized search on the hyperparameters of the classifier with the real dataset to maximize the balanced accuracy \cite{brodersen2010balanced}. The final set of hyperparameters was then used to re-train the classifier on the proxy data using the same procedures as before. This was done for each of the 4 sets of vital signs. The synthetic pre-release corresponds to the row in bold. Further details as well as PyTorch \cite{paszke2017automatic} implementations can be found in the repository \footnote{\anon{\url{https://github.com/3778/Ward2ICU}}}.

\paragraph{Protecting Commercial Sensitive Information} Classification tasks with imbalanced classes commonly report metrics that are a function $f$ of the confusion matrix $\mM$, such as $F_1$ score and Balanced Accuracy. However, $f\left(\mM\right) = f\left(\mM^\prime\right)$ does not imply that $\mM = \mM^\prime$, making it difficult to evaluate with $f$ if the GAN has properly learned to synthesize each individual class. We can not make a verbatim report on $\mM$, nor can we divulge multiple values of $f\left(\mM\right)$ for different $\mM$, as it would indirectly disclose \anon{HMD's} ICU to discharge ratio (ratio between positive and negatives classes). Hence, we opted to show the minority and majority class accuracies.\footnote{Note that we do not train nor cross-validate on these metrics, they are used solely for empirical evaluations.} To permanently hide this information, the proxy dataset was generated with balanced classes.


\begin{table}
    \centering
    \captionsetup{skip=0.5\baselineskip}
    \caption{GAN architecture parameterized by number of signals ($s$), label embedding ($c$), seed ($m$) and hidden layer ($h$) sizes. LeakyReLU (LReLU) activations \cite{xu2015empirical}, Replicated Padding (RP) and Dropout (DP) \cite{srivastava2014dropout} are also shown. All convolutional (Conv) layers have kernel size of 3 and stride 1. Upsampling uses linear interpolation and label embeddings (AppEmb) append along the last dimension.}
    \label{gan-network}
    \begin{tabular}{lcllcl}
        \toprule
        & Discriminator &&& Generator & \\
        \\
        \textbf{Layer} & \textbf{Act./Padd./Reg.} & \textbf{Output shape} & \textbf{Layer} & \textbf{Act./Padd./Reg.} & \textbf{Output shape}\\
        \cmidrule(r){1-3}
        \cmidrule(r){4-6}
        Input   &          & $20 \times s      $ & Seed     &             & $m$                \\
        AppEmb  &          & $20 \times (s + c)$ & AppEmb   &             & $m \times (1 + c)$ \\
                &          &                     & Linear   & LReLU/DP    & $\ 5 \times h$     \\
        Conv    & LReLU/RP & $20 \times h$       & Upsample &             & $10 \times h$      \\
        Conv    & LReLU/RP & $20 \times h$       & Conv     & LReLU/RP    & $10 \times h$      \\
        AvgPool &          & $10 \times h$       & Conv     & LReLU/RP/DP & $10 \times h$      \\
        Conv    & LReLU/RP & $10 \times h$       & Upsample &             & $20 \times h$      \\
        Conv    & LReLU/RP & $10 \times h$       & Conv     & LReLU/RP    & $20 \times h$      \\
        AvgPool &          & $\ \ 5 \times h$    & Conv     & LReLU/RP/DP & $20 \times h$      \\
        Linear  & LReLU/RP & $\ \ 1$             & Conv     &             & $20 \times s$      \\
        \bottomrule
    \end{tabular}
\end{table}

\section{Conclusion and Future Work}
We used a Conditional WGAN-GP to synthesize a proxy database of vital signs with an associated binary class label indicating patient transitions from the general ward to the ICU. Commercial sensitive information was hidden by balancing the generated dataset. Evaluation was done by comparing individual class accuracies for an LSTM classifier on proxy and original data. From our preliminary results in \autoref{results} we argue that data utility is being transferred from the original to the proxy dataset for the \emph{Ward2ICU} binary classification task. Some accuracy on the minority class is lost, as was expected. 

Future work will focus on circumventing issues that currently harm data publishing for and beyond patient privacy. Specifically, developing new ways to explicitly trade data utility for protection of commercial sensitive information and finding new ways to generate multi-modal EHR. One path to explore is applying the same theory used to obtain privacy guarantees for patients, called Differential Privacy \cite{dwork2006calibrating}, to commercial sensitive information.

\newpage
\hide{
    \subsubsection*{Acknowledgments}
    We would like to acknowledge the entire 3778 Research team for insightful discussions and reviews, especially Marcio Aldred Gregory for helping us define a relevant classification task and for bringing up the discussion on commercial sensitive data. We would also like to thank \anon{Hospital Mater Dei} for providing the original data. Finally, we would like to thank the founders of 3778 Healthcare, Guilherme Salgado and Fernando Barreto, for intensively investing in research in a country where there is little to no incentives for this type of work.
}
\printbibliography
\end{document}